\pdfoutput=1
\documentclass[journal]{IEEEtran}
\usepackage{epstopdf}
\ifCLASSINFOpdf
   \usepackage[pdftex]{graphicx}
   \graphicspath{{C:../pdf/}{C:../jpeg/}}
   \DeclareGraphicsExtensions{.pdf,.jpeg,.png}
\else
  \usepackage[dvips]{graphicx}
  \graphicspath{{C:../eps/}}
  \DeclareGraphicsExtensions{.eps}
\fi
\usepackage{amsmath}
\makeatletter

\newcommand{\Rmnum}[1]{\expandafter\@slowromancap\romannumeral #1@}
\makeatother
\usepackage{bm}
\usepackage{colortbl}
\usepackage{makecell}
\usepackage{multirow}
\usepackage{amssymb}
\usepackage[colorlinks, linkcolor=black, anchorcolor=black, citecolor=black]{hyperref}
\usepackage{threeparttable}
\usepackage{array}
\usepackage{float}
\usepackage{amsmath}

\usepackage{color}
\usepackage{cite}
\usepackage[ruled]{algorithm2e}
\usepackage{algpseudocode}
\usepackage{graphics}
\usepackage{epsfig}

\UseRawInputEncoding

\begin{document}

\title{Coordinating CAV Swarms at Intersections with a Deep Learning Model}%

\author{Jiawei Zhang, Shen Li, Li Li,~\IEEEmembership{Fellow,~IEEE}
\thanks{(\emph{Corresponding author: Li Li})}
\thanks{J. Zhang is with the Department of Automation, Tsinghua University, Beijing 100084, China.}
\thanks{S. Li is with the Department of Civil Engineering, Tsinghua University, Beijing 100084, China.}
\thanks{L. Li is with the Department of Automation, BNRist, Tsinghua University,
Beijing 100084, China (e-mail: li-li@tsinghua.edu.cn).}}

\markboth{arXiv}%
{Shell \MakeLowercase{\textit{et al.}}: Bare Demo of IEEEtran.cls for IEEE Journals}

\maketitle
\begin{abstract}
    Connected and automated vehicles (CAVs) are viewed as a special kind of robots that have the potential to significantly improve the safety and efficiency of traffic. In contrast to many swarm robotics studies that are demonstrated in labs by employing a small number of robots, CAV studies aims to achieve cooperative driving of unceasing robot swarm flows. However, how to get the optimal passing order of such robot swarm flows even for a signal-free intersection is an NP-hard problem (specifically, enumerating based algorithm takes days to find the optimal solution to a 20-CAV scenario). Here, we introduce a novel cooperative driving algorithm (AlphaOrder) that combines offline deep learning and online tree searching to find a near-optimal passing order in real-time. AlphaOrder builds a pointer network model from solved scenarios and generates near-optimal passing orders instantaneously for new scenarios. Furthermore, our approach provides a general approach to managing preemptive resource sharing between swarm robotics (e.g., scheduling multiple automated guided vehicles (AGVs) and unmanned aerial vehicles (UAVs) at conflicting areas).
\end{abstract}

\begin{IEEEkeywords}
Connected and automated vehicles (CAVs), cooperative driving, signal-free intersection, robot swarms, deep learning.
\end{IEEEkeywords}
\IEEEpeerreviewmaketitle
\section{Introduction}

\IEEEPARstart{C}{onnected} and automated vehicle (CAV) swarms are expected to be among the first large-scale robotic systems to enter and widely affect existing social systems, and to be the key participants in next-generation intelligent transportation system \cite{xu2019cooperative, wu2021flow, zhang2022analysis}. With the aid of vehicle-to-vehicle communication, CAVs can share their states (e.g., position, velocity, etc.) and intentions (e.g., behaviors, path, etc.) to achieve cooperative driving, improving efficiency while ensuring safety.

\par The major tasks of CAV swarms are pretty different from those of other kinds of mobile swarm robots \cite{dorigo2020reflections, mcguire2019minimal}. First, instead of exploring the unknown environment, CAV swarms are often assumed to be fully aware of driving environments. Second, the communication between swarms is fast and reliable \cite{talamali2021less}; the guidance message can be instantly delivered from the roadside decision support unit to swarms. Third, we do not require CAV swarms to obtain certain formations \cite{berlinger2021implicit}. Generally, we aim to guide every swarm to run along the road as quickly as possible and pass the conflicting area without collisions. Fourth, unlike many swarm robotics studies focused on a small number of robots, CAV swarm studies targeted unceasing robot swarm flows that arrive at conflicting regions. Usually, we try to turn conventional signalized intersections to be signal-free so that the traffic efficiency can be boosted by better organizing the passing order of CAV swarms. The number of CAVs in swarm flows is infinitely large, and the instantaneous arrivals at a certain conflict area is also potentially large.

\par In such conditions, the core problem is to determine the sequential order of CAVs to occupy a certain conflicting area \cite{xu2019cooperative, reveliotis2011conflict}. One typical example for such right-of-way arrangements is to find the optimal passing order that can minimize the total delay of all CAVs for a signal-free intersection. It has been shown in papers \cite{meng2017analysis, yu2021automated, zhang2022analysis} that a better passing order can improve the performance of the CAV swarms system, while a poor order may cause traffic congestion.

\par It is extremely challenging to find the (near-)optimal passing order because: (\emph{i}) the number of passing orders grows exponentially with the number of vehicles, e.g., there are approximately $8.159 \times 10^{47}$ passing orders for the scenario with 40 vehicles; (\emph{ii}) algorithm is generally deployed at a roadside unit and the edge computing capability is limited; (\emph{iii}) algorithm should satisfy the real-time requirement since vehicles are moving.

\par Researchers have proposed various algorithms for solving passing orders in the last two decades. These algorithms can be roughly categorized into three kinds. The first category is the reservation algorithms that roughly follow the first-come-first-served (FCFS) principle \cite{chai2018connected, mitrovic2019combined, medina2019optimal}. In each round of planning, the arrival time of CAVs to the conflict area will be estimated, and then the passing order is arranged in ascending order of their arrival times. Although the FCFS based algorithms have a low computational burden, its generated passing order is usually far from the optimal solution \cite{meng2017analysis, zhang2022analysis, levin2016optimizing}. The second category is mathematical programming algorithms which transfer the whole problem into an optimization problem with several integer variables introduced to denote the relative order between vehicles \cite{zhu2015linear, fayazi2018mixed, lu2019trajectory, hult2020optimisation}. However, due to the combination explosion, the resulting mixed-integer programming problem is often intractable to solve when the size of CAV swarms is large (more than 20 CAVs).

\begin{figure*}[t]
\centering
\includegraphics[width=6.1in]{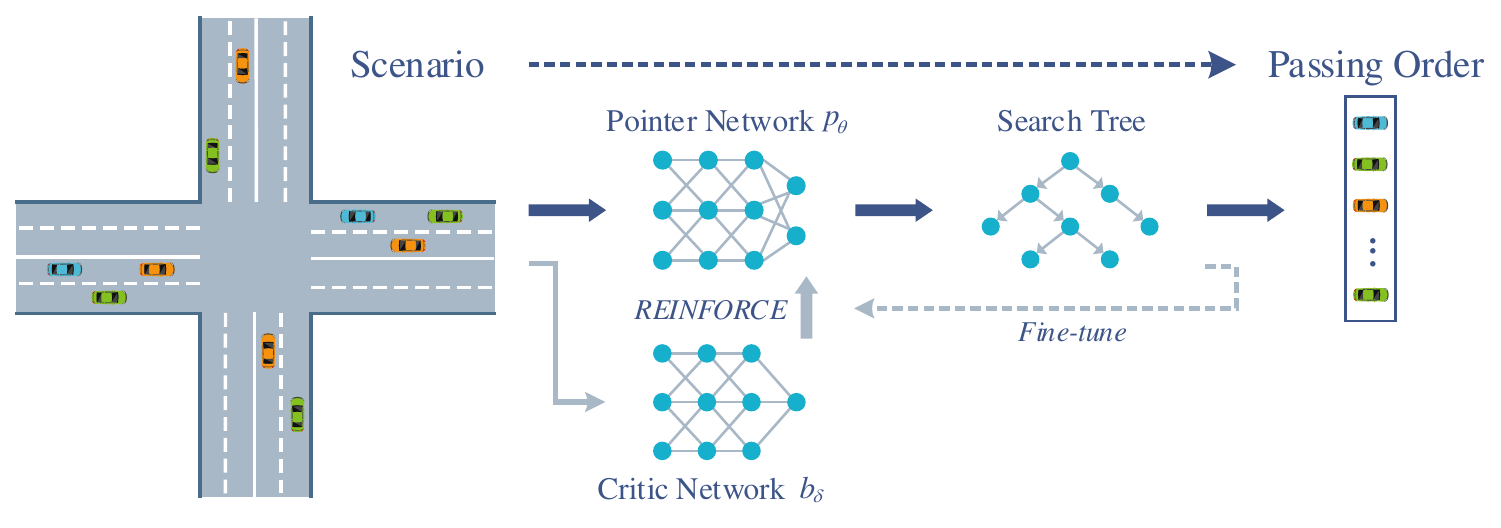}
\caption{ Training and online solving pipeline of AlphaOrder. AlphaOrder takes vehicles within the intersection scenario as the input, and its output is the passing order of these vehicles passing through the intersection. It has two core modules, where the pointer network $p_{\theta}$ is used to solve candidate passing order, and the search tree is used to carry out further searching. Critic network $b_{\delta}$ works as a baseline to assist the REINFORCE algorithm to train the pointer network $p_{\theta}$. Besides, the solutions improved by search tree can also be used to fine-tune the pointer network $p_{\theta}$ further.}
\label{fig1}
\end{figure*}

\par The third category is tree search algorithms that describe the passing order as a sequence of symbols (CAVs) and formulate the problem as a tree search problem \cite{li2013survey, xu2020bi, zhang2021bi}. The detailed tree formulations are diverse and usually dependent on the search algorithms. For example, starting from an empty root node, the child node can be expanded by adding an uncovered vehicle's symbol at the end of the partial order denoted by its parent node. This formulation makes the Monte Carlo tree search (MCTS) algorithm easy to apply \cite{xu2019cooperative}. Existing tree search algorithms do not exhaustively search the whole tree but use some handcrafted heuristics to narrow the search to a beam of promising branches. However, the performance of handcrafted heuristics highly depends on the number of vehicles and may degrade with the increase of vehicles.

\par Here we present AlphaOrder, a new algorithm that combines deep learning and tree searching to achieve state-of-the-art performance in solving passing order for unceasing CAV swarm flows. The main idea of the algorithm is to employ neural networks to learn the patterns of (near-)optimal passing orders for the solved scenarios and instantaneously generate good enough candidate passing orders for new scenarios. To achieve this, we formulate the problem as a combinatorial optimization problem and use a pointer network \cite{vinyals2015pointer} (an excellent neural network architecture for solving sequence-to-sequence (Seq2Seq) problems) to capture the underlying relation between a scenario and its promising passing orders. Based on the candidate passing order, we can further employ a very-short-time tree search to possibly upgrade the solution and meanwhile keep the planning time short enough for online applications.

\par We summary the developed AlphaOrder algorithm in Fig. \ref{fig1}. We first train a pointer network $p_{\theta}$  using the well-known REINFORCE algorithm \cite{bello2016neural} and meanwhile employs a critic network $b_{\delta}$ as the baseline for variance reduction. We choose REINFORCE algorithm to train the pointer network $p_{\theta}$ mainly because supervised learning (SL) is not applicable. The optimal solution for our problem can only be solved by exhaustive enumeration, resulting in that obtaining high-quality labels for a large number of training scenarios requires intolerable computation time. Instead, policy based reinforcement learning (RL) provides an unsupervised paradigm for training the pointer network. It has been shown that RL can automatically discover underlying experiences on some challenging problems, e.g., the traveling salesman problem (TSP) \cite{bello2016neural}. After training, we use the pointer network $p_{\theta}$ to produce candidate passing orders. Subsequently, we formulate a search tree based on the candidate passing order and employ the MCTS algorithm to search within a limited time budget (0.1s). The results demonstrate that AlphaOrder can find near-optimal passing orders for scenarios with an arbitrary number of vehicles. Specifically, for a problem instance containing 40 vehicles, the passing order solved by AlphaOrder in 0.14s is comparable to the best passing order that can be found by partly enumerating 50 days.

\par To better present our work, the rest of this paper is organized as follows. \emph{Section \ref{sec2}} gives a concise problem description. \emph{Section \ref{sec3}} introduces the proposed AlphaOrder algorithm. In \emph{Section \ref{sec4}}, we show numerical experiments to demonstrate its performance. \emph{Section \ref{sec5}} concludes this paper. Below, for reading convenience, the main symbols and their definitions used in this paper are listed in APPENDIX Table \ref{table_appendix_1}.

\section{cooperative driving as a combinatorial optimization problem}
\label{sec2}

\par In order to solve with pointer network and MCTS, we formulate the underlying problem into a combinatorial optimization problem whose solution, i.e., passing order, is denoted by a sequence. Its solution space and objective function are formulated in detail as followings:

\textbf{\emph{Solution Space.}} We use a sequence of the permutation of vehicles' symbols to concisely denote passing orders. For example, the sequence \emph{``AB''} means that \emph{Vehicle} \emph{A} is given priority over \emph{Vehicle} \emph{B} in the right-of-way arrangement. For a scenario \emph{s} containing \emph{N} vehicles $\{ V_1, V_2, \cdots, V_{N} \}$, a complete passing order can be denoted by a sequence \bm{$\pi$} $= \{\pi_1, \pi_2, \cdots, \pi_{N} \}$, where each $\pi_{k}$ $(k=1,2, \cdots, N)$ is the symbol of any vehicle being placed at the \emph{k}-th position of \bm{$\pi$}. All possible passing orders form a solution space \bm{$\Pi$} consisting of $N!$ sequences.

\begin{figure*}
\centering
\includegraphics[width=6.5in]{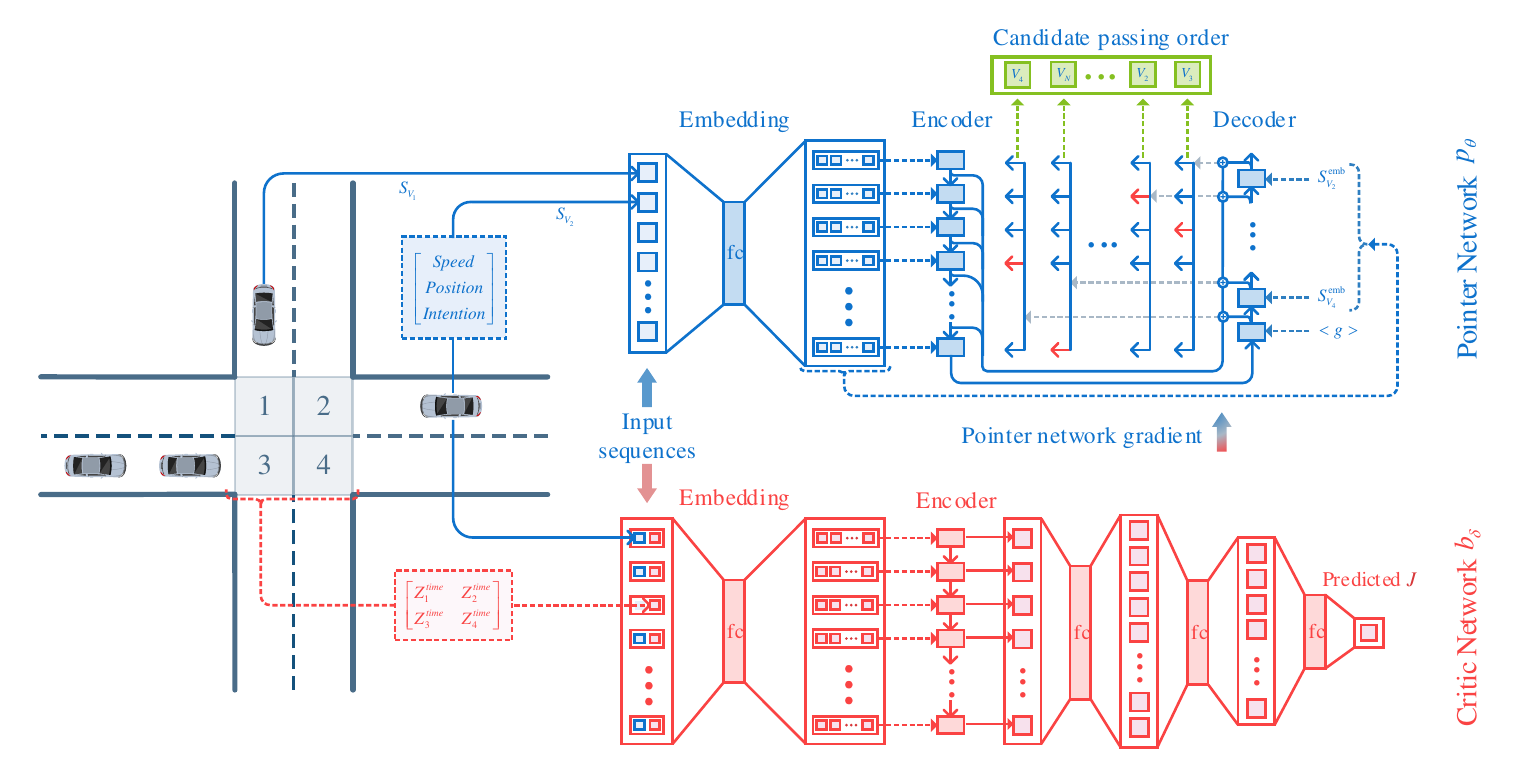}
\caption{Pointer and critic network architecture. (\emph{i}) Pointer network $p_{\theta}$: A linear embedding layer first embeds state information $S_{V_i} (i=1,2,\cdots, N)$ consisting of speed, position, and intention (steering and route) for each vehicle. Then, the sequence consisting of the vehicles' embedding information $\{ S^{emb}_{V_i} \in \mathbb{R}^{D_{emb}}\}$ is fed into a modified encoder-decoder model, which consists of two independent LSTM cells. In each decoding step, the pointer network $p_{\theta}$ calculates the conditional probability $p_{\theta}(\pi_k \mid \cdot)$ over unselected vehicles (denoted by $\oplus$), and selects the vehicle with the highest probability (indicated by the red arrow) to be added to the passing order $\bm{\pi}$; (\emph{ii}) Critic network $b_{\delta}$: The embedding layer and encoder of the critic network $b_{\delta}$ have the same architecture as the pointer network $p_{\theta}$. The encoded information is calculated by multilayer fully connected networks to output a prediction of objective value $J$. $<g>$ is the input for the first step of the decoder and is a \emph{D}-dimensional learnable vector; $Z^{time}_i (i=1,\cdots,4)$ is right-of-way state of conflict area \cite{xu2019cooperative}; fc: fully connected layer.}
\label{fig2}
\end{figure*}

\textbf{\emph{Objective Function.}} We aim to minimize the total delay by rearranging the passing order. Therefore, the objective function of the combinatorial optimization problem is formulated as
\begin{equation}
\min J(\bm{\pi}\mid s)=\sum_{i=1}^{N} J_{\text{Delay}}\left(V_{i}, \bm{\pi}\mid s\right) + C \times f_{\text {Enforceable}}(\bm{\pi} \mid s)
\label{equ1}
\end{equation}
Here, $J_{\text{Delay}}\left(V_{i}, \bm{\pi}\mid s\right)$ is the delay of \emph{Vehicle i} with the passing order $\bm{\pi} \in \bm{\Pi}$, and thus the first term is the delay-sum of all vehicles. Besides, since it is unenforceable for passing orders that violate the order of vehicles in front and behind at the same lane, the second bool function $f_{\text {Enforceable}}(\bm{\pi} \mid s)$ is used to guarantee the enforceability of $\bm{\pi}$. An appropriate penalty factor $C$ is set to guide the training of the neural networks to avoid unenforceable solutions.

\section{ALPHAORDER ALGORITHM}
\label{sec3}

\subsection{Pointer Network Module}
\label{sec3-A}

\par Because all vehicles $\{ V_1, V_2, \cdots, V_{N} \}$  will be arranged in passing order $\bm{\pi} \in \bm{\Pi}$ sequentially, we can naturally convert the above combinatorial optimization problem into a Seq2Seq problem. The input sequence is the state information of all vehicles, denoted by  $\bm{S} = \{S_{V_1}, S_{V_2}, \cdots, S_{V_N}\}$. The output sequence is a permutation of the elements in $\bm{S}$, i.e., the passing order $\bm{\pi}$  in our problem, which aims to minimize the objective value $J$ in equation (\ref{equ1}).

\par Here, we employ a pointer network $p_{\theta}$ parameterized by $\theta$ to solve this Seq2Seq problem since the pointer network has been proved to be effective in combinatorial optimization problems \cite{bello2016neural}. The pointer network is a particular network architecture capable of learning the conditional probability of the output sequence, where the elements are indexes corresponding to the positions in the input sequence. For a given input sequence $\bm{S} = \{S_{V_1}, S_{V_2}, \cdots, S_{V_N}\}$, the pointer network assigns an occurrence probability $p(\bm{\pi} \mid \bm{S})$ to each possible passing order $\bm{\pi} \in \bm{\Pi}$ and calculate it as a product of conditional probabilities according to the chain rule
\begin{equation}
p(\bm{\pi} \mid \bm{S})=\prod_{k=1}^{N} p_{\theta}\left(\pi_{k} \mid \pi_{1}, \cdots, \pi_{k-1}, \bm{S}\right)
\label{equ2}
\end{equation}
Here, the production of a passing order $\bm{\pi}$ is decomposed into an \emph{N}-step vehicle selection procedure, where the \emph{k}-th ($k=1,2,\cdots,N$) step selects one vehicle from the input sequence $\bm{S}$ and places it at $\pi_k$. The conditional probability $p_{\theta}(\pi_k \mid \cdot)$ models the probability of any vehicle being selected at step \emph{k} according to the given $\bm{S}$ and vehicles that have been selected \cite{bello2018seq2slate}. A well-trained pointer network $p_{\theta}$  can assign high probabilities to passing orders with small $J$ and low probabilities to passing orders with large $J$ .

\par The general neural networks for solving Seq2Seq problems encode the input sequence into a vector that contains the input information by a recurrent neural network (RNN) and subsequently decode the vector into the output sequence by another RNN. Usually, we named these two RNNs as ``encoder" and ``decoder" \cite{zhu2021uav}. Similarly, as depicted in Fig. \ref{fig2}, our pointer network $p_{\theta}$ employs two independent Long Short-Term Memory (LSTM) cells (a class of RNN frequently used at learning long-term dependent features) to serve as encoder and decoder, respectively. The decoder incorporates the attention mechanism as a pointer to select vehicles from input sequence as the output. The designed encoder and decoder are elaborated as follows:

\par \textbf{\emph{Encoder.}} The encoder network is used to obtain a representation for each vehicle in the input sequence $\bm{S}$. In order to extract useful features from $\bm{S}$ more efficiently, we first linearly embed each element in $\bm{S}$ into a high $D_{emb}$-dimension vector space, obtaining $\bm{S}^{emb} = \{ S^{emb}_{V_1},S^{emb}_{V_2}, \cdots, S^{emb}_{V_N}  \}$ where $S^{emb}_{V_i} \in \mathbb{R}^{D_{emb}} (i=1,\cdots, N)$. Then, the embedding vectors $\bm{S}^{emb}$ are fed into the first LSTM cell. At each encoding step $i$ $(i=1,\cdots, N)$, the first LSTM cell reads one vector $S^{emb}_{V_i}$ and outputs a  $D$-dimensional latent memory state $e_{V_i} \in \mathbb{R}^D$. After $N$ steps of encoding, we obtain a latent representation sequence $\bm{e}=\{e_{V_1}, e_{V_2}, \cdots, e_{V_N}\}$.

\par \textbf{\emph{Decoder.}} The decoder network is used to produce the output sequence, i.e., the passing order $\bm{\pi}$, iteratively by \emph{N} steps of decoding. At each decoding step $k$ $(k=1,2,\cdots,N)$, the latent memory state $d_k \in \mathbb{R}^D$ containing information from previous decoding steps is output by using the second LSTM cell. Subsequently, the decoder employs the attention mechanism to calculate the conditional probability $p_{\theta}(\pi_k \mid \cdot)$ based on $d_k$ and the encoded representation sequence $\bm{e}=\{e_{V_1}, e_{V_2}, \cdots, e_{V_N}\}$. Here, the attention mechanism can give different weights to different vehicles of the input, thus extracting the latent relationship between each vehicle and the already output partial passing order $\{ \pi_1, \pi_2, \cdots, \pi_{k-1}\}$ at the current decoding step \emph{k}. Finally, the decoder selects the vehicle with the maximum conditional probability as $\pi_k$, and passes the vehicle as the input to the next decoding step. Specifically, the conditional probability $p_{\theta}(\pi_k \mid \cdot)$ is calculated by
\begin{equation}
u_{i}^{k}=\left\{\begin{split}
& v^{T} \tanh \left(W_{1} e_{V_{i}}+W_{2} d_{k}\right), \text {if } V_i \notin \left\{\pi_{1}, \cdots, \pi_{k-1}\right\} \\
&~~~~~~~~~~~ -\infty,~~~~~~~~~~~~~\text {Otherwise }~~~~
\end{split}\right.
\label{equ3}
\end{equation}
\begin{equation}
p_{\theta}\left(\pi_{k} \mid \pi_{1}, \cdots, \pi_{k-1}, \bm{S}\right)=\operatorname{softmax}\left(u^{k}\right)
\label{equ4}
\end{equation}
where $v \in \mathbb{R}^{D \times 1}$, $W_1$, $W_2 \in \mathbb{R}^{D \times D}$ are the learnable attention parameters and are collectively denoted by $\theta$. The softmax function normalizes the vector $u^k$ to a probability distribution over the unselected vehicles. Essentially, $u^k_i$ denotes the degree to which \emph{Vehicle i} is placed at $\pi_k$.

\par In order to output a promising passing order, we need to find the optimal model parameters $\theta^*$ for the pointer network $p_{\theta}$ from problem instances. Here, we choose the REINFORCE algorithm to optimize $\theta$, as it has proved to be an appropriate paradigm for training neural networks to solve combinatorial optimization problems \cite{bello2016neural, nazari2018reinforcement}. Its core idea is to iteratively update $\theta$ by stochastic gradient descent in the direction that minimizes the objective value $J$. Meanwhile, to reduce the variance of the gradients during the training phase, a critic network $b_{\delta}$ parameterized by $\delta$ is used to provide an approximated baseline of objective value $J$ for any problem instance \cite{konda1999actor}.  Specifically, given a scenario $s$ and a passing order $\bm{\pi}$, and the gradient of $\theta$ is formulated by REINFORCE algorithm:
\begin{equation}
\Delta \theta \propto \frac{\partial \log p_{\theta}(\boldsymbol{\pi} \mid \boldsymbol{S})}{\partial \theta}\left(J(\bm{\pi} \mid s)-b_{\delta}\left(\boldsymbol{S}^{\prime}\right)\right)
\label{equ5}
\end{equation}
where $\boldsymbol{S}^{\prime}$ is the input sequence for critic network $b_{\delta}$. Each element in $\boldsymbol{S}^{\prime}$ contains the vehicle state $S_{V_i}$ and also concatenates the intersection right-of-way state to improve the prediction performance. We train the critic network $b_{\delta}$ by using stochastic gradient descent to minimize the mean square error (MSE) between the predicted objective value $b_{\delta} (\boldsymbol{S}^{\prime})$ and the actual objective value $J(\bm{\pi} \mid s)$, and thus the gradient of $\delta$ can be calculated by
\begin{equation}
\Delta \delta \propto \frac{\partial b_{\delta}\left(\boldsymbol{S}^{\prime}\right)}{\partial \delta}\left(J(\bm{\pi} \mid s)-b_{\delta}\left(\boldsymbol{S}^{\prime}\right)\right)
\label{equ6}
\end{equation}

\par The critic network $b_{\delta}$ works as an auxiliary network to provide an approximated baseline of objective value $J$. As depicted in red in Fig. 2, our critic network $b_{\delta}$ has the same architecture as the encoder of the pointer network $p_\theta$. The encoded latent state representation is passed through three fully connected layers with size $d_{fc,1}$, $d_{fc,2}$, $d_{fc,3}$ to produce a prediction of objective value $J$. Here, since the right-of-way states of conflict area have a slight impact on the objective value $J$, we also input them into the critic network $b_{\delta}$ for improved performance (illustrated as $Z^{time}_i (i=1,\cdots,4)$ in Fig. 2).

\begin{algorithm}[h]
\label{alg1}
\caption{Pointer and Critic Network Training.}
\LinesNumbered
\KwIn{Training dataset $\Psi$, batch size $B$.}
\KwOut{$\theta$, $\delta$.}
Initialize pointer network weights $\theta$ and critic network weights $\delta$ with random weights.\\
\While{weights not converge}{
Sample $s_i \sim \Psi$, $i=1,2,\cdots,B$\\
Calculate $\bm{\pi}_i(i=1,2,\cdots, B)$ with pointer network $p_\theta$.\\
Calculate $b_{\delta}(\bm{{S}}^{\prime}_i)(i=1,2,\cdots, B)$ with critic network $b_\delta$.\\
Calculate the gradient of $\theta$:
$$
\Delta \theta \leftarrow \frac{1}{B} \sum_{i=1}^{B} \frac{\partial \log p_{\theta}\left(\boldsymbol{\pi}_{i} \mid \boldsymbol{S}_{i}\right)}{\partial \theta}\left(J\left(\boldsymbol{\pi}_{i} \mid s_{i}\right)-b_{\delta}\left(\boldsymbol{S}_{i}^{\prime}\right)\right)
$$\\
Calculate the gradient of $\delta$:
$$
\Delta \delta \leftarrow \frac{1}{B} \sum_{i=1}^{B} \frac{\partial b_{\delta}\left(\boldsymbol{S}_{i}^{\prime}\right)}{\partial \delta}\left(J\left(\boldsymbol{\pi}_{i} \mid s_{i}\right)-b_{\delta}\left(\boldsymbol{S}_{i}^{\prime}\right)\right)
$$\\
Update $\theta \leftarrow \operatorname{ADAM}(\theta, \Delta \theta)$.\\

Update $\delta \leftarrow \operatorname{ADAM}(\delta, \Delta \delta)$.\\
}
\end{algorithm}

\begin{figure*}
\centering
\includegraphics[width=6.5in]{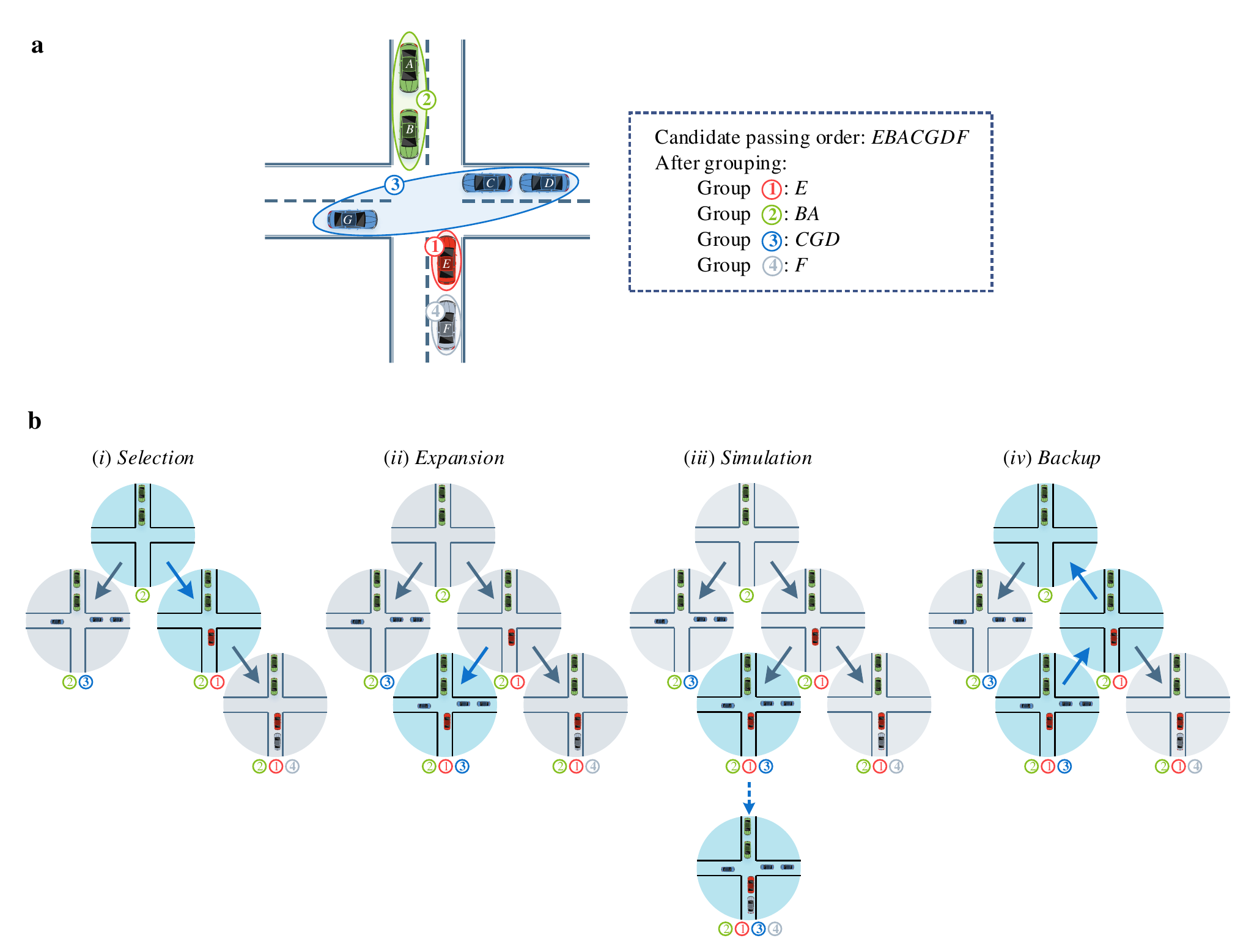}
\caption{Grouping and Monte Carlo tree search in AlphaOrder. \textbf{a}, Grouping: Vehicles that have adjacent positions in candidate passing order are bound together into one group. The search tree grows with groups as the basic units (instead of vehicles). \textbf{b}, MCTS algorithm: (\emph{i}) \emph{Selection}: selecting the most promising nodes based on their scores; (\emph{ii}) \emph{Expansion}: expanding the not-yet-visited child node into the current search tree; (\emph{iii}) \emph{Simulation}: running several rollout simulations until a leaf node is reached, i.e., a complete passing order is obtained; (\emph{iv}) \emph{Backup}: back-propagating to update the scores of all parent nodes based on the score of the leaf node.}
\label{fig3}
\end{figure*}

\par Algorithm \ref{alg1} presents the training procedure of the pointer network $p_{\theta}$ and the critic network $b_{\delta}$, which has a similar training framework as the asynchronous advantage actor-critic (A3C) algorithm \cite{mnih2016asynchronous}. In Algorithm \ref{alg1}, the weights $\theta$ and $\delta$ are updated asynchronously and iteratively. In each iteration, $B$ problem instances $s_i(i=1,2,\cdots,B)$  are first sampled from the training dataset $\Psi$ (line 3) and fed into the pointer network $p_\theta$ (line 4) and critic network $b_\delta$ (line 5), respectively. The corresponding passing order $\bm{\pi}_i$ and predicted objective value $b_{\delta}(\bm{{S}}^{\prime}_i)$ are calculated through forward propagation. Then, the gradient of $\theta$ is calculated by equation (\ref{equ5}) (line 6), and the gradient of $\delta$ is calculated by equation (\ref{equ6}) (line 7). Finally, we adopt the ADAM optimizer \cite{kingma2014adam} to update the weights via back propagation (line 8 and 9). This completes one iteration, and the algorithm can be terminated when the weights converge, i.e., the mean objective value $J$ on the training dataset converges. Moreover, for each type of vehicle number \emph{N}, we construct a dataset $\Psi$ containing 200,000 problem instances to train the corresponding pointer network $p_{\theta}$ and the critic network $b_{\delta}$. All the networks can be well-trained within 48\emph{h}. Once trained, the candidate passing orders produced by the pointer network $p_{\theta}$ for new scenarios have substantially outperformed the best existing algorithms (detailed results shown in Fig. \ref{fig9}).


\subsection{Tree Search Module}
\label{sec3-B}

\par AlphaOrder uses a short-time tree search to further improve the optimality of the passing order solved for online applications. Here, unlike AlphaGo \cite{silver2016mastering}, we need not exploit the learned experiences by using the neural networks directly to guide the tree search. This is because the passing order is a cooperation between vehicles rather than a game. Moreover, the online applications do not have the time budget to allow us to use neural networks to guide extensive tree search. To this end, we develop an indirect and more efficient way of incorporating the learned experiences into the tree formulation and then employing the MCTS algorithm to traverse the tree.

\par The tree formulation in AlphaOrder builds on previous work on cooperative driving using some handcrafted heuristics \cite{xu2019grouping}. Noticing that to consider vehicles in groups may reduce calculation time, we bind the conflict-free vehicles that have adjacent positions in candidate passing order as one group to formulate the search tree, which is then used as the basic unit for the tree formulation. Supposing that all vehicles are divided into $M$ groups, our search tree grows as follows: its root node is empty; each node in the first level contains one group; the second level adds another group on the first-level nodes, and so on until all groups are added. Hence, each node at level $M$, i.e., the leaf node, will correspond to a complete passing order containing all groups. As illustrated in Fig. \ref{fig3}a, the relative order of vehicles in one group is fixed during searching. Compared to the original tree with vehicles as the basic units, our tree formulation approach significantly reduces the depth and width of the search tree.

\par We employ the MCTS algorithm to search the tree formulated from candidate passing order. MCTS algorithm is a best-first search algorithm based on sampling as a node evaluation \cite{williams1992simple, coulom2006efficient}. Instead of relying on heuristic domain knowledge, it uses Monte Carlo rollouts to narrow down the search to promising branches. Its effectiveness has been demonstrated by the outstanding performance in searching huge solution space, such as Go \cite{silver2016mastering} and Chess \cite{schrittwieser2020mastering}. Each node in the search tree is assigned a score to evaluate its potential. The score is determined by the mean objective value $J$ over the subtree below that node and the visit count. The scores of nodes are continuously updated during the tree traversal. MCTS algorithm selects the search direction based on the scores of the nodes. Fig. \ref{fig3}b illustrates the searching procedure of MCTS, which iteratively traverses the tree, with each iteration comprising four steps \cite{chaslot2008monte}: \emph{selection, expansion, simulation, and backup}. As more simulations are executed, the tree is traversed broader and deeper, and the current optimal passing order will be gradually updated. The four steps are elaborated as follows.

\subsubsection{Selection} Starting from the root node, MCTS selects promising child nodes to search forward according to a selection strategy. The key of the selection strategy is to balance the exploitation of nodes with high scores and the exploration of unvisited nodes. Here, we use the classical UCB1 \cite{kocsis2006bandit,baier2014mcts} selection strategy. Although there are various modified selection strategies, we employ UCB1 because of its simplicity and effectiveness. Moreover, without any prior knowledge regarding the search tree, UCB1 can promisingly address the exploration-exploitation dilemma in MCTS \cite{browne2012survey}. UCB1 calculates a score based on the accumulated visit count and value of the child node and then selects the search direction, as follows:
\begin{equation}
\underset{i}{\arg \max } Q_{i}+\lambda \sqrt{\frac{\ln T}{T_{i}}}
\label{equ7}
\end{equation}
where $Q_{i}$ is the value of the child node \emph{i}. $T_i$ is the accumulated visit count of child node \emph{i}. $T$ is the accumulated visit count of the current node. $\lambda$ is a mixing parameter. In the UCB1 formula (\ref{equ7}), the first term is for exploitation, and the second term is for exploration. $\lambda$ trades off between the two. The child node with the highest score is selected layer by layer, up to an expandable node. An expandable node is the one that has unvisited child nodes.

\subsubsection{Expansion} When an expandable node is reached with the selection strategy, one or more of its child nodes will be expanded to the tree. Here, we randomly select one of the unvisited child nodes and expand it to the tree.

\subsubsection{Simulation} Staring from the expanded node, several rollout simulations are performed to reach a leaf node to obtain a complete passing order, whose delay-sum will be used to evaluate the value of the expanded node. The classical MCTS randomly selects one group per rollout simulation that has not been added yet and goes deeper without divergence until a leaf node is reached. However, similarly, due to the physical constraint of the front and behind of vehicles in the same lane, random selection potentially may make the reached leaf node unenforceable \cite{xu2019cooperative}. Here, we prefer to add groups close to the conflict area to make the leaf node executable. We use  $\bar{J_i}$ to denote the delay-sum of the partial passing order corresponding to the expanded node and $\hat{J_i}$ to denote the delay-sum of the leaf node, normalized by:
\begin{equation}
q_{i}=1-\frac{J_{i}-J_{i, \min }}{J_{i, \max }-J_{i, \min }}
\label{equ8}
\end{equation}
where $J_{i, \max }$ and $J_{i, \min }$ are the maximum and minimum delay-sums of the sibling nodes of the node \emph{i}. Then, the value $Q_i$ for the expanded node is:
\begin{equation}
Q_{i}=\gamma \bar{q}_{i}+(1-\gamma) \hat{q}_{i}
\label{equ9}
\end{equation}
where $\gamma$ is a weighting parameter to balance the potential of the node and its subtree.

\subsubsection{Backup} The value of the expanded node is successively back-propagated through the search path to update the values of all parent nodes.

\par Depending on the corresponding applications, the above four steps are repeated a fixed number of loops or until the time budget runs out. In this paper, due to real-time requirements, we set the search time budget to 0.1s. When the predefined search time budget runs out, AlphaOrder outputs the leaf node with the shortest delay-sum as the best passing order.

\section{Numerical Experiments}
\label{sec4}
\begin{figure}[h]
\centering
\includegraphics[width=3.5in]{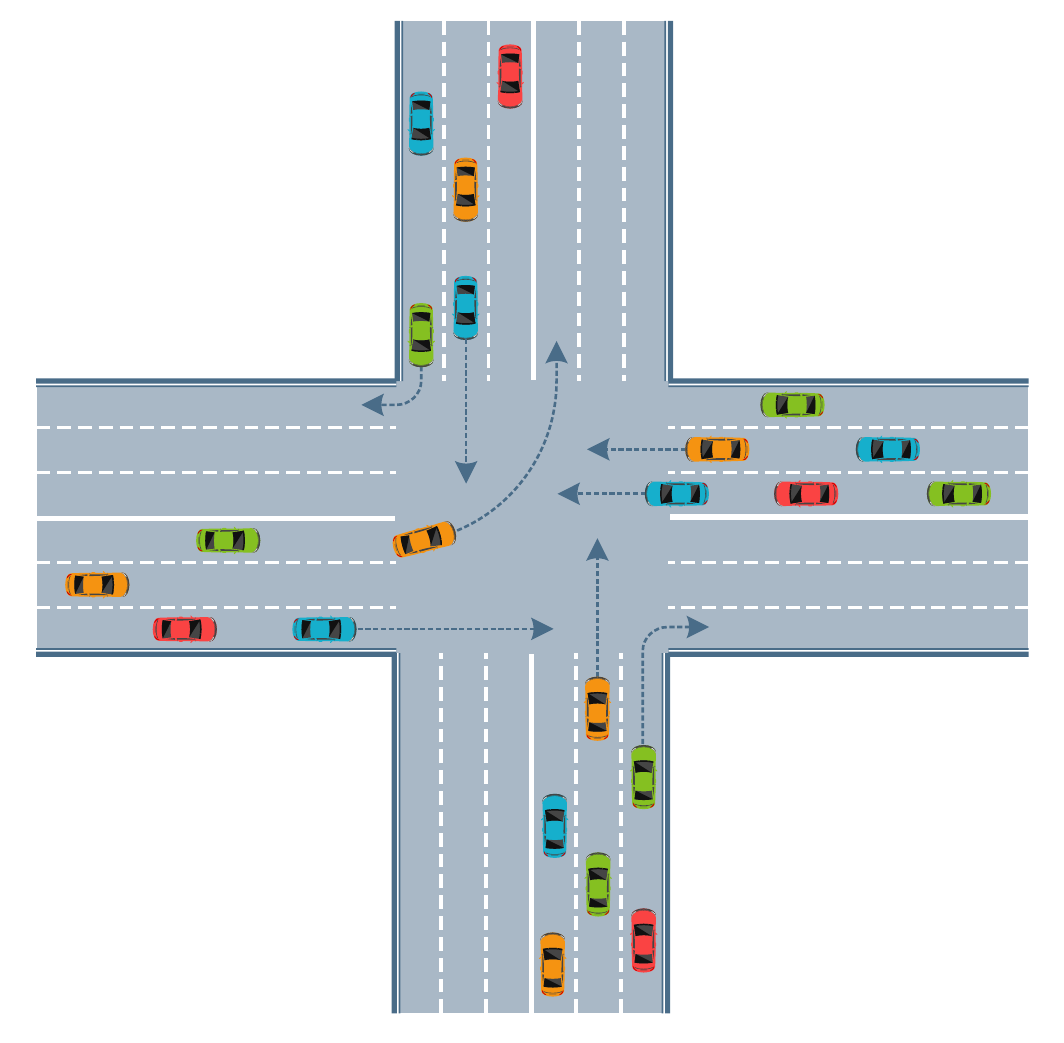}
\caption{A typical signal-free intersection with three lanes in each direction.}
\label{fig4}
\end{figure}

In this section, we will take a three-lane intersection as an example to demonstrate the performance of the AlphaOrder algorithm (as illustrated in Fig. \ref{fig4}). \emph{Section \ref{sec4-a}} describes the experiments settings. \emph{Section \ref{sec4-b}}  discusses the choice of key parameters. \emph{Section \ref{sec4-b}} evaluates the performance of AlphaOrder algorithm from three aspects. In \emph{Section \ref{sec4-d}} ,  we verify the generalization performance of AlphaOrder for different traffic conditions.

\subsection{Experiments Settings}
\label{sec4-a}

\subsubsection{State representation}
\par We take the three-lane intersection to illustrate the specific state representation of the input sequences $\bm{S}$ and $\boldsymbol{S}^{\prime}$ for pointer network $p_\theta$ and critic network $b_\delta$.  The state information $S_{V_i}$ of \emph{Vehicle i} $(i=1,2, \cdots,N)$  is characterized by four parts: the longitudinal speed, the distance to the conflict area, the steering type, and the route. First, the longitudinal speed and the distance to the conflict area are one-dimensional scalars and will be normalized before being fed into the neural networks. Second, for vehicle steering, we use 0, 1, and 2 to characterize left-turn, straight, and right-turn, respectively. Finally, since there are 12 entry lanes, we encode them with a 12-dimensional one-hot vector, with the vehicle's lane set to 1 and the rest to 0. The exit lane is also encoded with a 12-dimensional one-hot vector. The two vectors are concatenated to represent the route information. Thus, the element in the input sequence $\bm{S}$ of pointer network $p_\theta$  will be an 27-dimensional vector. Besides, the conflict area is divided into 36 conflict subzones, and thus the right-of-way state of the conflict area can be denoted by a 36-dimensional vector $\{Z^{time}_1,Z^{time}_2, \cdots, Z^{time}_{36}\}$. Here, $Z^{time}_i$ is the assignable time (i.e., right-of-way) of the corresponding conflict subzone \emph{i} \cite{xu2019grouping}.  Thus, the element in the input sequence $\boldsymbol{S}^{\prime}$ will be a 63-dimensional vector. Moreover, it should be noted that without loss of generality, we arrange the vehicles' position in the input sequences $\bm{S}$ and $\boldsymbol{S}^{\prime}$ according to an ascending order of their distances from the conflict area.

\subsubsection{Neural network and training settings}
\par For the example three-lane intersection, the specific size of the neural networks is set in a general way. For the pointer network $p_\theta$, the dimension $D_{emb}$ of the linear embedding layer is 256. The dimension $D$ of all latent memory states is 256. For the critic network $b_\delta$, the size of the embedding layer and the encoder is the same as that of the pointer network $p_\theta$. The size of the three subsequent fully connected layers are $d_{fc,1}=1024$, $d_{fc,2}=256$, and $d_{fc,3}=1$, where we use rectified linear activation functions \cite{glorot2011deep} in the first two fully connected layers.

\par A batch size $B$ of 512 problem instances was used to train the neural networks. The initial learning rate for the ADAM optimizer is 0.001. We designed a custom learning rate scheduler, where the learning rate was kept for the first 10000 training iterations, and then the learning rate was multiplied by 0.98 at every 1000 iterations. In order to comply with the practical computing power of the roadside unit, the following training and evaluations are all conducted on a single machine consisting of one Inter i7 CUP and one NVIDIA GeForce RTX 3080 GPU.

\subsubsection{Simulation of signal-free intersections}
The signal-free intersections studied in this paper are mainly simulated on the cooperative driving simulation platform presented in paper \cite{zhang2022analysis}.With the unceasing arriving of new CAV swarms, AlphaOrder solves the cooperative driving problems in a rolling planning manner, which can cope with an infinite number of CAVs. Furthermore, it is worth noting that deriving the objective value $J$ in equation (\ref{equ1}) for a given passing order is convenient, with time complexity of $O(n)$ \cite{xu2019grouping}.

\subsection{The Choice of Parameters}
\label{sec4-b}

\begin{figure}[h]
\centering
\includegraphics[width=3.5in]{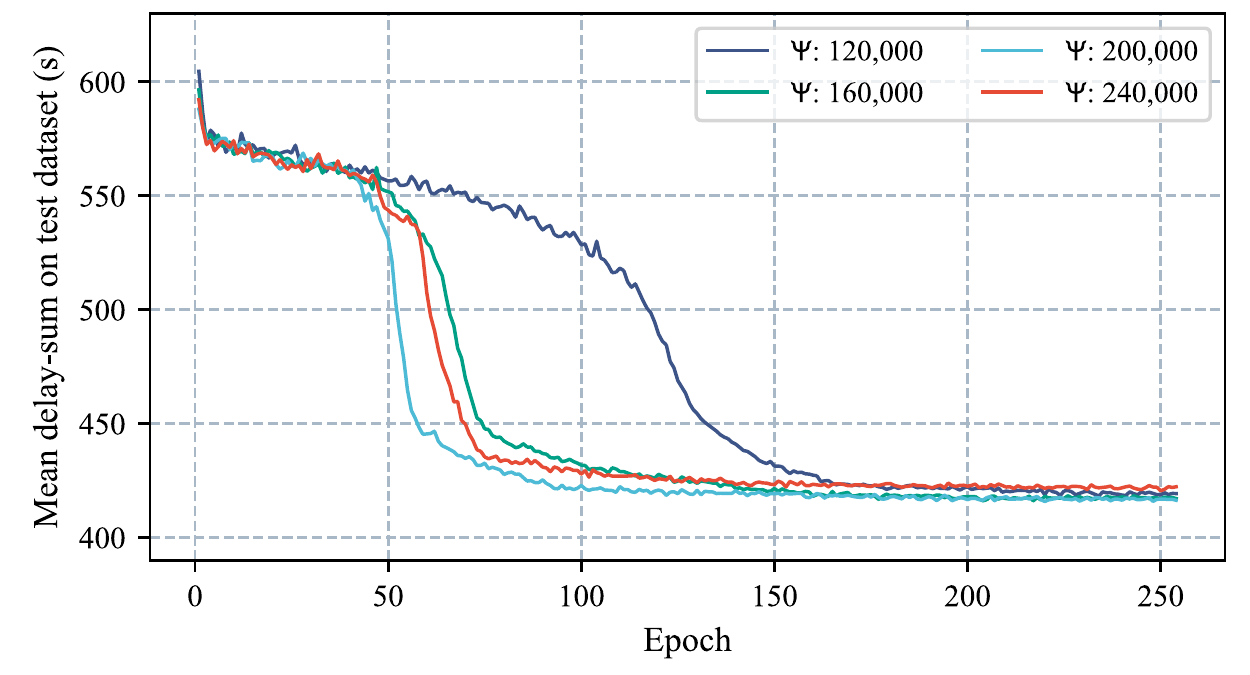}
\caption{Effects of training dataset size.We construct four different training datasets containing 120000, 160000, 200000, and 240000 instances, respectively. Our models are trained on these four datasets and are evaluated on the same test dataset (1000 independent problem instances) every one epoch after the first epoch. The number of vehicles $N=30$.}
\label{fig5}
\end{figure}

\subsubsection{Training dataset}
\par To train a neural network with excellent performance, we must sample enough problem instances uniformly over a distribution covering all possible scenarios to build the training dataset $\Psi$. Our approach for achieving this is simple, i.e., we sample online problem instances directly from the simulated signal-free intersection. For the size of the training dataset $\Psi$, the results shown in Fig. \ref{fig5} suggest that the final performance of the neural network is robust to different dataset sizes, with only minor differences in the training speed and stability. Even on a small training dataset ($\Psi$ : 120,000), AlphaOrder achieves roughly the same performance after a little more exploration, demonstrating AlphaOrder¡¯s excellent learning ability. Among them, a dataset of size 200,000 is a promising and general choice for balancing the time consuming and model performance. Moreover, although the same pointer network $p_\theta$ and critic network $b_\delta$ can process scenarios with different numbers of vehicles, we differentiate the dataset according to the number of vehicles and train them separately. We take this approach because the delay-sums and the underlying patterns of the promising passing orders for scenarios with different vehicles vary greatly, which is detrimental to the training of the neural networks.


\begin{figure}[h]
\centering
\includegraphics[width=3.5in]{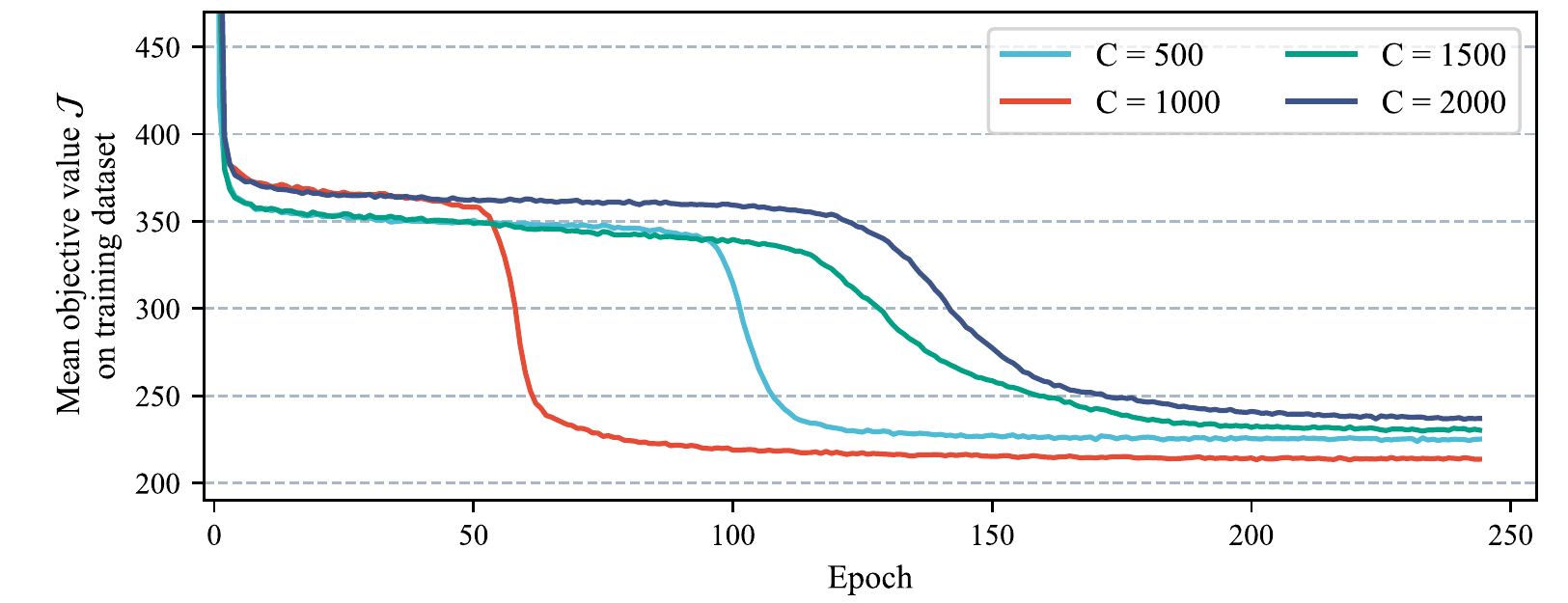}
\caption{Effects of hyper-parameter $C$ on the training of neural networks.}
\label{fig6}
\end{figure}

\subsubsection{Penalty factor $C$}
 A critical hyper-parameter affecting the training is the penalty factor $C$ in equation (\ref{equ1}). The value of $C$ needs to be appropriate. If it is too small, it will not play the role of penalty, and if it is too large, it will result in a slow learning speed. Here, to find a suitable value, we set $C$ to 500, 1000, 1500, and 2000, respectively. Fig. \ref{fig6} shows the training of neural networks with different $C$. It suggests that the pointer network $p_\theta$ has a faster learning speed and shorter objective value $J$ when $C$ is 1000. Thereby we set $C$ to 1000 when training the neural network. Moreover, it should be noted that the mean objective value $J$ can be reduced to a reasonable stage after only one epoch, which is precisely caused by the penalty term to avoid unenforceable passing orders.

\begin{figure}[h]
\centering
\includegraphics[width=3.5in]{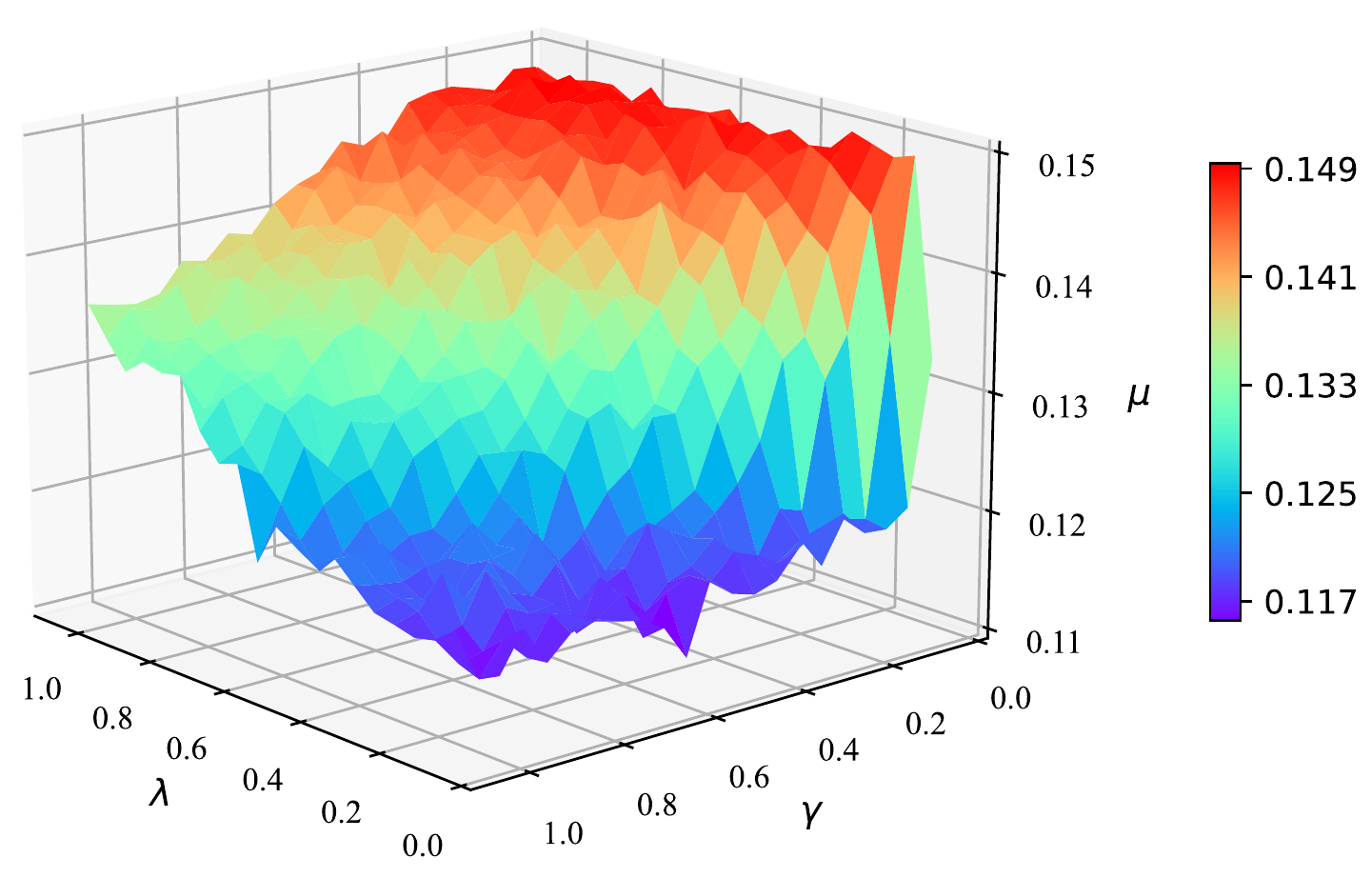}
\caption{Effects of tree search hyper-parameters on the performance of AlphaOrder.}
\label{fig7}
\end{figure}

\subsubsection{Hyper-parameter selection for MCTS}
MCTS has two key hyper-parameters to be addressed, i.e., $\lambda$ in equation (\ref{equ7}) and $\gamma$ in equation (\ref{equ9}). We carried out a grid search for them to choose a better $\lambda$ and $\gamma$. The values of $\lambda$ and $\gamma$ vary between 0 and 1 with step 0.05, and we evaluate different settings on 1000 problem instances, respectively. We define the following improvement ratio $\mu$ to demonstrate the performance of different settings:
\begin{equation}
\mu=\frac{J_{\text {CandidatePassingOrder }}-J_{\text {AlphaOrder }}}{J_{\text {CandidatePasssingOrder }}}
\label{equ10}
\end{equation}
where $J_{\text {CandidatePassingOrder }}$ is the delay-sum of the candidate passing order, and $J_{\text {AlphaOrder }}$ is the delay-sum of the passing order found by tree search. Fig. \ref{fig7} shows the average improvement ratio $\mu$ with different $\lambda$ and $\gamma$. The result suggests that the tree search module can improve the optimality of the passing order under arbitrary parameter settings. The result also shows that as  $\lambda$ increases and $\gamma$ decreases, the improvement ratio $\mu$ increases, i.e., the delay-sum of the searched passing order decrease. Since $\lambda=0.85$ and $\gamma=0.15$ achieve the maximum $\mu=0.1503$, we use them in other experiments.

\subsection{Evaluation of AlphaOrder}
\label{sec4-c}
Here, we implement two other typical cooperative driving algorithms for comparative experiments. One is the First-In-First-Out based reservation algorithm (abbreviated as FIFO based algorithm), which determines an enforceable passing order according to the order of vehicles entering the intersection control area and has the advantages of low computational burden and good real-time performance. The other is excellent tree search algorithm that combines MCTS with some heuristic rules \cite{xu2019cooperative} (abbreviated as MCTS based algorithm). In the following, we will evaluate the AlphaOrder algorithm from three aspects. First, we evaluate the superiority of AlphaOrder by comparison; second, we evaluate the learning speed of AlphaOrder; and finally, we demonstrate the efficient deployment of AlphaOrder to new arbitrary intersections by transfer learning.

\begin{figure}[h]
\centering
\includegraphics[width=3.5in]{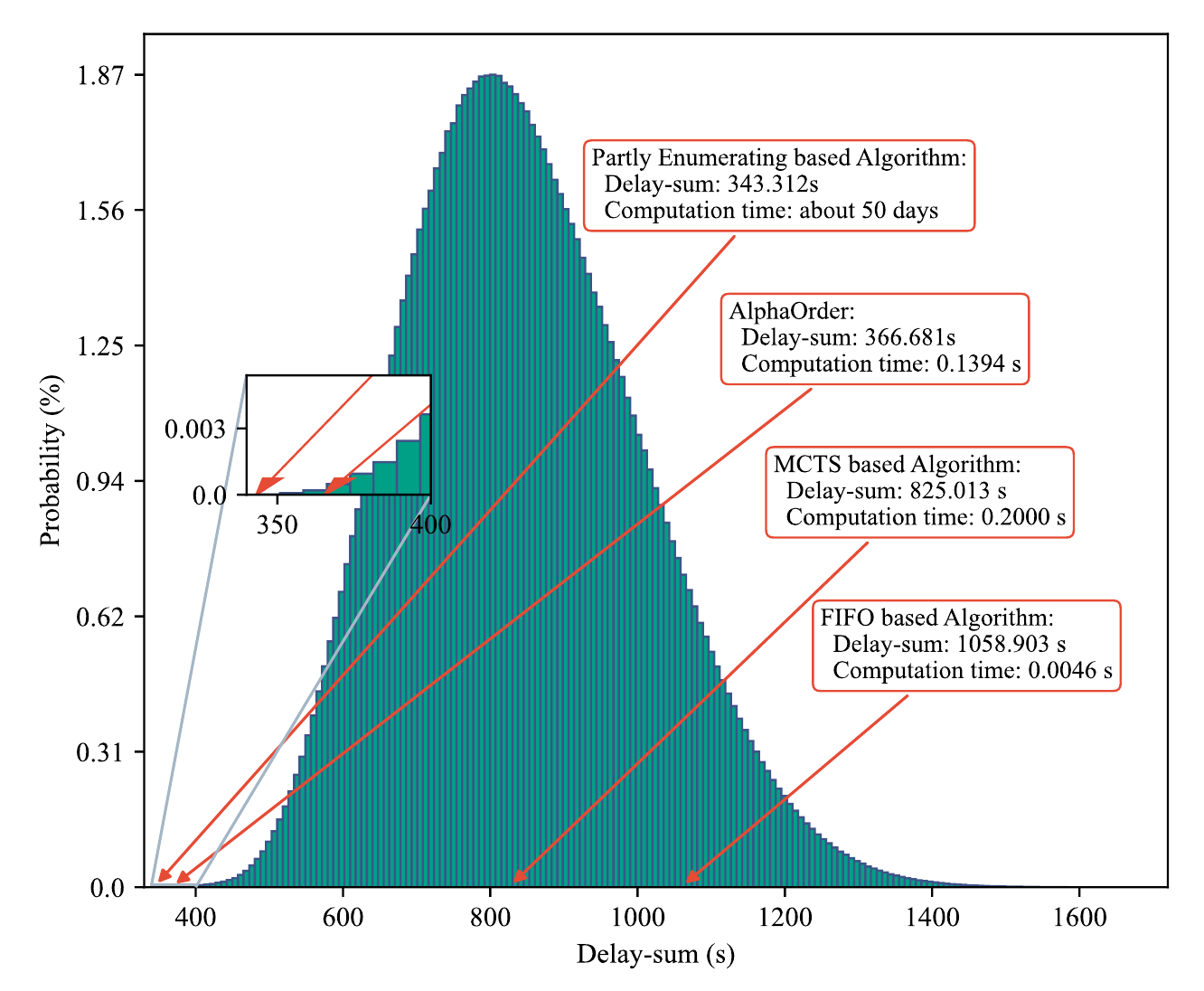}
\caption{The histogram of delays of different passing orders for a problem instance with 40 vehicles.}
\label{fig8}
\end{figure}

\begin{figure}[h]
\centering
\includegraphics[width=3.5in]{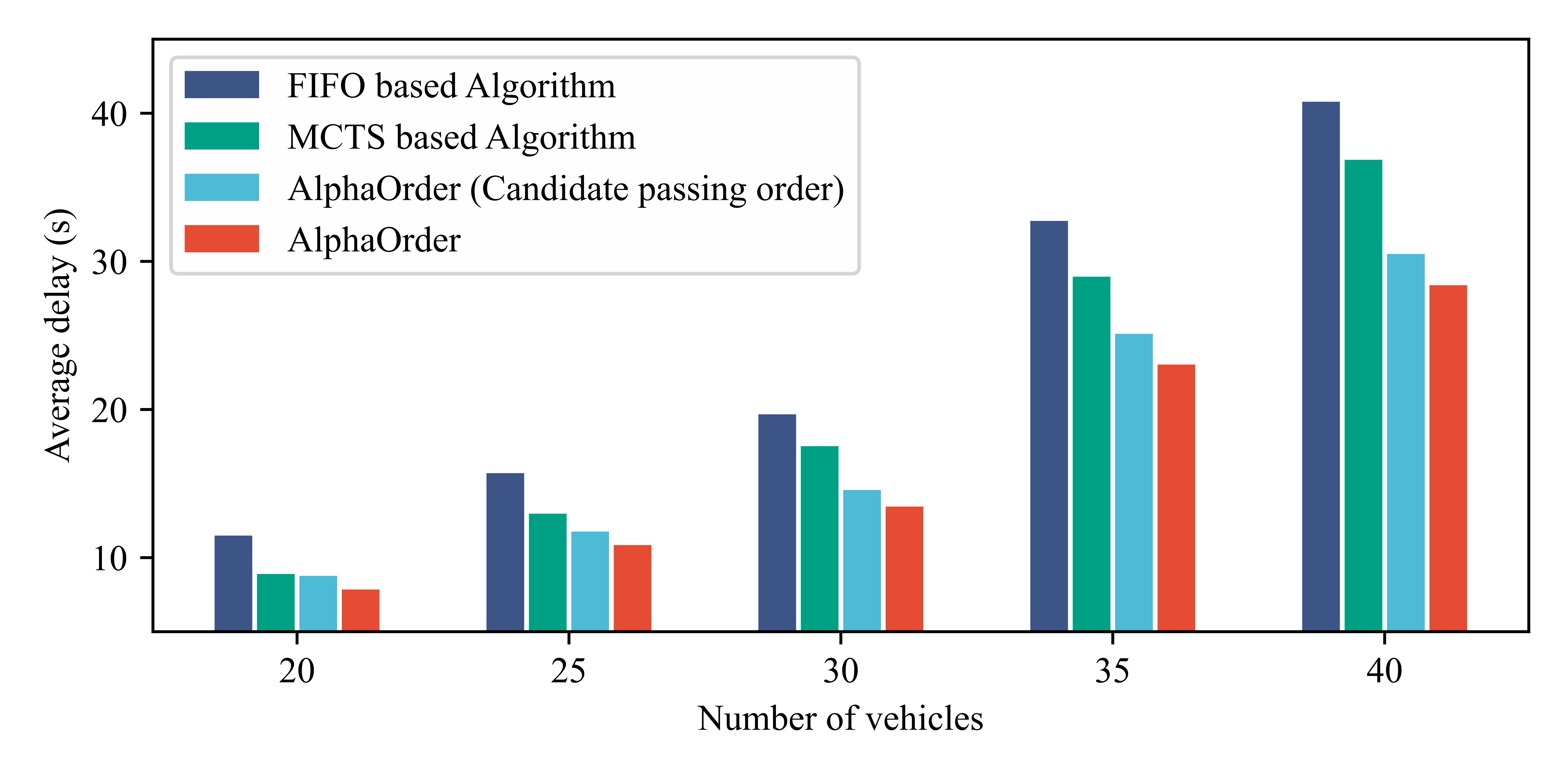}
\caption{The performance of AlphaOrder versus other algorithms on scenarios with different numbers of vehicles. To facilitate comparison, we count the average delay of vehicles, i.e., $\sum_{i=1}^{N} J_{\text {Dealy }}\left(V_{i}\right) / N$ where $\sum_{i=1}^{N} J_{\text {Dealy }}\left(V_{i}\right) $ is the delay-sum and $N$ is the number of vehicles.}
\label{fig9}
\end{figure}

\subsubsection{Superiority of AlphaOrder}
To clearly show the performance of the AlphaOrder algorithm, we study a problem instance with 40 vehicles in detail. Since enumerating all the passing orders is intolerably time-consuming, we spent about 50 days partly enumerating about 60 million enforceable passing orders uniformly in its solution space. In other words, we uniformly sample over the solution space. Inspired by our tree formulation approach, we designed a simple method to achieve this: grouping the vehicles randomly. Specifically, we randomly group vehicles within the same lane, and the front-to-behind constraints of the vehicles determine the partial order within the group. In this way, the enumeration with these groups as the basic units will enhance the reflection for the whole solution space.

\par We visualized them in a histogram manner (as shown in Fig. \ref{fig8}). It is clear that the passing order solved by AlphaOrder in a short time (the pointer network $p_\theta$ takes 0.04s and the subsequent tree search takes 0.1s) is roughly equal to the global optimal passing order found by partly enumerating and is located in a solution space with a small probability. Meanwhile, AlphaOrder is significantly outstanding than FIFO and MCTS based algorithms, and the corresponding delay-sum is reduced by $65.37\%$ and $55.55\%$, respectively.

\par Furthermore, to evaluate the average performance of the AlphaOrder algorithm, we build a test dataset containing 1000 independent instances for each vehicle number $N$. Fig. \ref{fig9} shows the average delay of the passing orders solved by different algorithms for $N=$20, 25, 30, 35, 40. To demonstrate the critical role of the two modules in AlphaOrder, we also plot the candidate passing order produced by the pointer network $p_\theta$. The results suggest that for scenarios containing an arbitrary number of vehicles, AlphaOrder can find the passing order with a shorter delay-sum, significantly outperforming the FIFO and MCTS based algorithms. The results also suggest that the pointer network $p_\theta$ can directly produce promising candidate passing orders for new scenarios, while the short-time tree search can further improve the optimality of the passing order. Furthermore, we can see that the MCTS based algorithm becomes increasingly inferior to our AlphaOrder as the number of vehicles increases, which is exactly caused by the exponential growth of the solution space.

\begin{figure}[htb]
\includegraphics[width=3.5in]{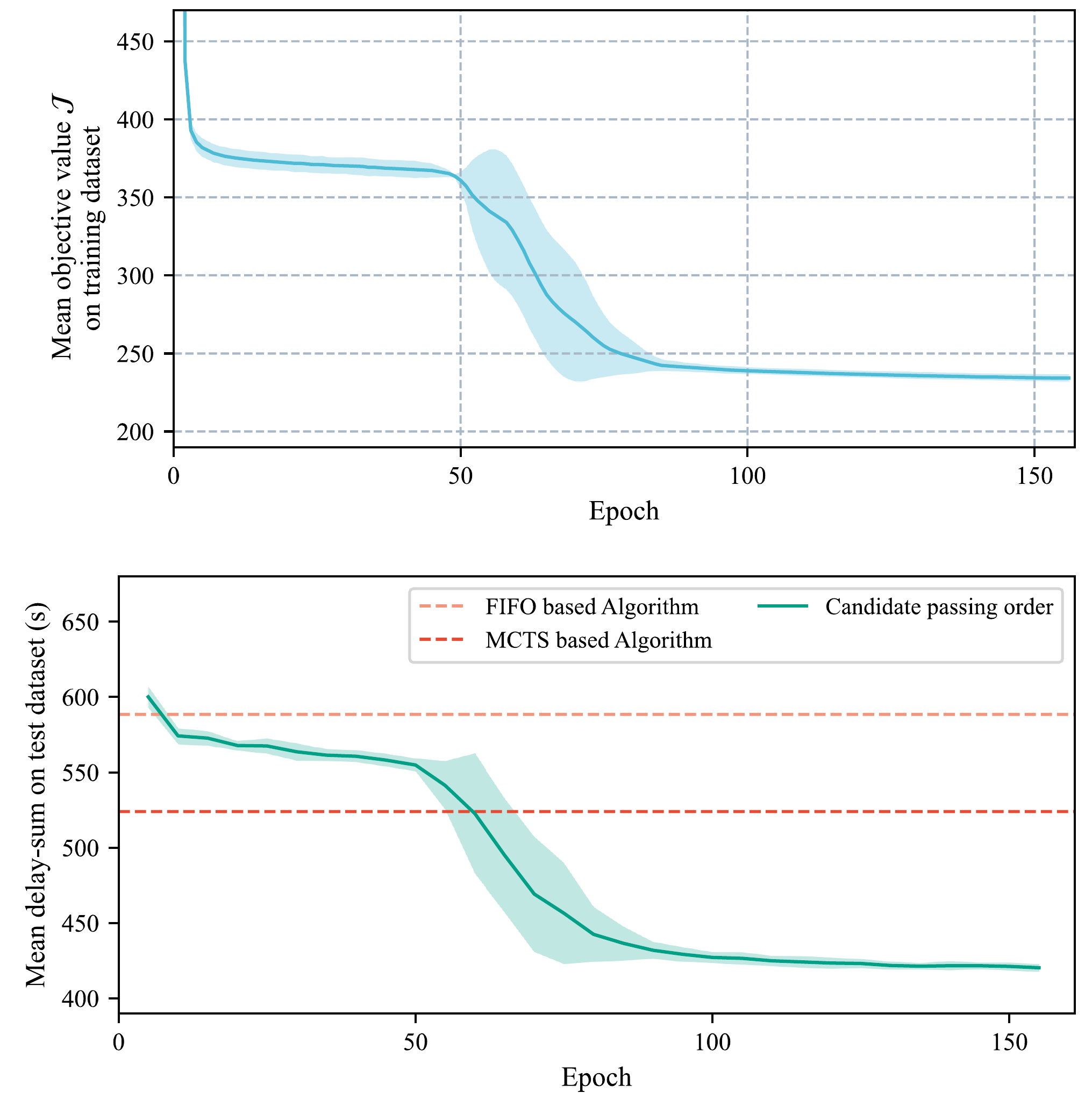}
\caption{Evaluation of AlphaOrder throughout training. (\textbf{\emph{Upper}})The mean objective function $J$. Results are averaged over five random seeds, where the solid line is the average, and the shading is the one standard deviation from the average. The horizontal axis is the training epoch, and each epoch corresponds to the neural network seeing once over all training instances. (\textbf{\emph{Lower}}) Performance of the pointer network $p_\theta$ corresponding to different epochs on the test dataset. We evaluate every five epochs after the first epoch. For comparison, we also plot the performance of MCTS and FIFO based algorithms. The number of vehicles  $N$=30. }
\label{fig10}
\end{figure}

\subsubsection{Learning speed}
To dissect the learning speed of the AlphaOrder algorithm, we evaluate the performance of pointer network $p_\theta$ throughout training. Fig. \ref{fig10} show the performance of pointer network $p_\theta$ on the training and test dataset as a function of the training epoch. Here, the training dataset contains 200,000 problem instances; similar to Fig. \ref{fig9}, the test dataset contains 1000 independent instances. The results show that pointe network $p_\theta$ can avoid the unenforceable passing orders after only one epoch, i.e., the front-to-behind constraints of vehicles in the same lane can be learned after seeing the training instances once. After about 50 epochs of exploration, the performance starts to improve rapidly. All the networks can be well trained within 150 epochs (taking about 48 hours on a single machine). Fig. \ref{fig10} also shows that after no more than 75 epochs, the candidate passing orders produced by pointer $p_\theta$ starts to outperform the MCTS based algorithm. Moreover, the results also show that the learning procedure of AlphaOrder has good stability.

\subsubsection{Efficient deployment to new intersections}
\par The proposed AlphaOrder can cope with arbitrary intersections. However, in practice, we do not want to take about 48 hours to train a model from scratch. Here, we will show that the experiences learned from one intersection can be quickly used to provide promising passing orders for arbitrary intersections by short-time transfer learning.

\begin{figure}[htb]
\includegraphics[width=3.5in]{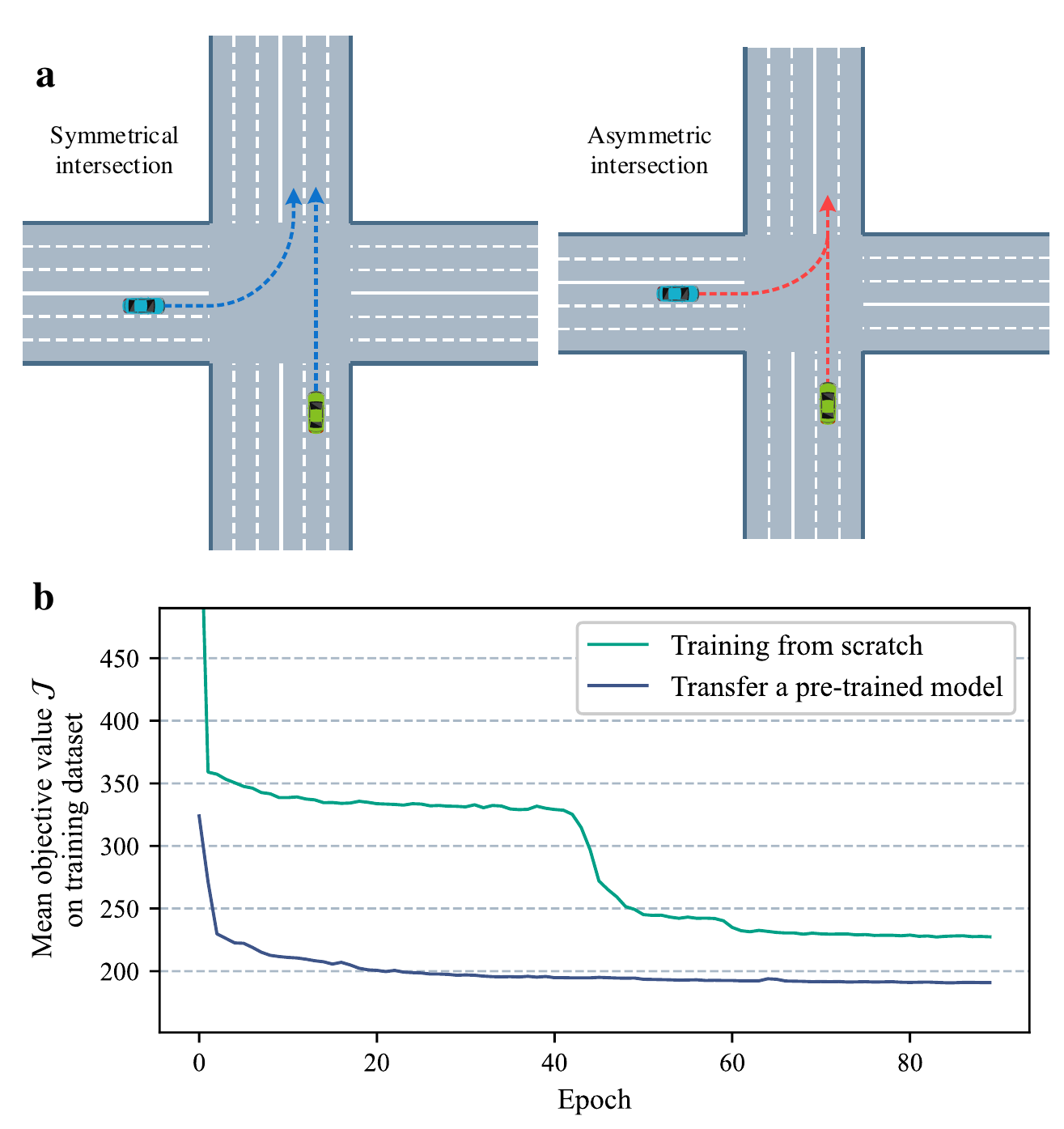}
\caption{Efficient deployment to new intersections. \textbf{a,} Intersections with symmetrical and asymmetric topologies. Vehicles that are conflict-free in the symmetric intersection encounter conflicts in the asymmetric intersection, indicating that the underlying patterns of promising passing orders differ from different intersections. \textbf{b,} The mean objective function   of training neural networks from scratch versus from the pre-trained model for the asymmetric intersection. The number of vehicles  $N$=30. }
\label{fig11}
\end{figure}

\begin{table*}[!htb]
\footnotesize
\renewcommand{\arraystretch}{1.3}
\centering
\caption{Average delay under different turning ratios and arriving rates.}
\centering
\setlength{\tabcolsep}{4.5mm}{
\begin{tabular}{cccccccc}
\hline\hline
\multirow{3}{*}{} && \multicolumn{6}{c}{Arriving rate (\emph{veh/(lane*h)})}              \\ \cline{3-8}
               Turning ratio   &Algorithm  & 200 & 220 & 240 & 260 & 280 & 300 \\ \cline{3-8}
                 &  & \multicolumn{6}{c}{Average delay (\emph{s})}              \\ \hline
\multirow{3}{*}{} &  FIFO based algorithm&     3.74 & 8.01  & 9.56  & 11.91 & 22.08 & 38.05     \\ \cline{2-8}
                  \emph{Left: 0.5; Right: 0.5} &MCTS based algorithm  &  2.33 & 6.03  & 7.06  & 9.51  & 19.68 & 33.92   \\ \cline{2-8}
                  & AlphaOrder &    2.32 & 5.25  & 6.01  & 7.59  & 13.10 & 20.79   \\ \hline
\multirow{3}{*}{} &  FIFO based algorithm&    7.87 & 15.17 & 25.50 & 31.05 & 29.88 & 37.33    \\ \cline{2-8}
                  \emph{Left: 0.8; Right: 0.5} &MCTS based algorithm  &    6.66 & 12.49 & 22.52 & 27.94 & 26.94 & 34.21      \\ \cline{2-8}
                  & AlphaOrder &   5.40 & 9.99  & 16.67 & 19.99 & 20.61 & 26.20    \\ \hline
\multirow{3}{*}{} &  FIFO based algorithm&    1.72 & 2.26  & 5.93  & 9.36  & 9.33  & 25.60  \\ \cline{2-8}
                  \emph{Left: 0.5; Right: 0.8} & MCTS based algorithm &  1.04 & 1.45  & 4.32  & 7.63  & 7.33  & 23.00  \\ \cline{2-8}
                  & AlphaOrder &     1.03 & 1.45  & 4.03  & 6.16  & 6.18  & 16.77  \\
\hline\hline
\end{tabular}}
\label{table_2}
\end{table*}

\par We can efficiently cope with such alteration by fine-tuning the experiences in the pre-trained model, i.e., transfer learning. Specifically, thanks to the fact that the design of our neural networks is intersection topology-agnostic, the internal architectures of the pointer network $p_\theta$ and the critic network $b_\delta$ need not be changed for new intersections. As depicted in Fig. 2, the only part that needs to be changed is the vector dimension of input sequence $\boldsymbol{S}^{\prime}$ for critic network $b_\delta$ , which varies with the specific conflict area. Thus, we can share the parameters  $\theta$ and $\delta$ of a pre-trained model except for the embedding layer of the critic $b_\delta$, and then fine-tune them by a short-time transfer training.

\par We demonstrate this with the intersections shown in Fig. \ref{fig11}a, where the pre-trained model comes from a symmetric intersection, and the transfer learning is done for the asymmetric intersection. Fig. \ref{fig11}b shows the mean objective value $J$ during training. The results show that the pre-trained model can be efficiently transferred to a new intersection, and the pre-trained model achieves a shorter $J$ for a short-time training than the model trained from scratch. Even without fine-tuning (corresponding to 0 epoch), the pre-trained model can find enforceable passing orders for the new intersections, suggesting a latent commonality of experiences between different intersections. Furthermore, the pre-trained model can roughly achieve the final performance after only 3 epochs (taking about 20min on a single machine). This is beneficial for practical large-scale deployments, i.e., given a pre-trained model, we can quickly deploy the model at diverse intersections by short-time transfer learning.

\subsection{Generalization to Different Traffic Conditions}
\label{sec4-d}

\par Turning ratio and traffic demand have a critical impact on cooperative driving \cite{zhang2022analysis}. Here, we evaluate the generalization capability of the AlphaOrder algorithm by varying the vehicle arriving rate, turning ratio, and traffic demand. It should be noted that problem instances used for training AlphaOrder are sampled in traffic with homogeneous traffic demand, i.e., the scenario distribution used for training is different from the following test settings. We deployed the cooperative driving algorithm on the simulated signal-free intersection and conducted simulation for 20min in each setting. After simulating, we count the average delay of vehicles passing through the intersection.

\par Table \ref{table_2} shows the average delay of vehicles under different turning ratios and arriving rates. The results show that AlphaOrder can achieve the shortest average delay under all turning ratios, demonstrating its excellent generalization performance for turning ratios. Moreover, with the increases of vehicle arriving rate, i.e., essentially with the increases of vehicles, AlphaOrder becomes more prominent in shortening travel delay. Besides, the average delay varies with different turning ratio settings. The reason for this phenomenon is that different turning types occupy distinct intersection right-of-ways, resulting in diverse vehicle conflictions. For example, left-turning and straight-going vehicles will prevent vehicles in other directions from passing, forming a complex conflict relation. On the contrary, right-turning vehicles will occupy less right-of-way and lead to fewer conflicts. Therefore, with the increases of the left-turning and straight-going vehicles, the conflicts among vehicles becomes more complex.  In essence, the results show that our AlphaOrder algorithm has outstanding generalization performance for different conflict relations.

\begin{figure}[h]
\includegraphics[width=3.5in]{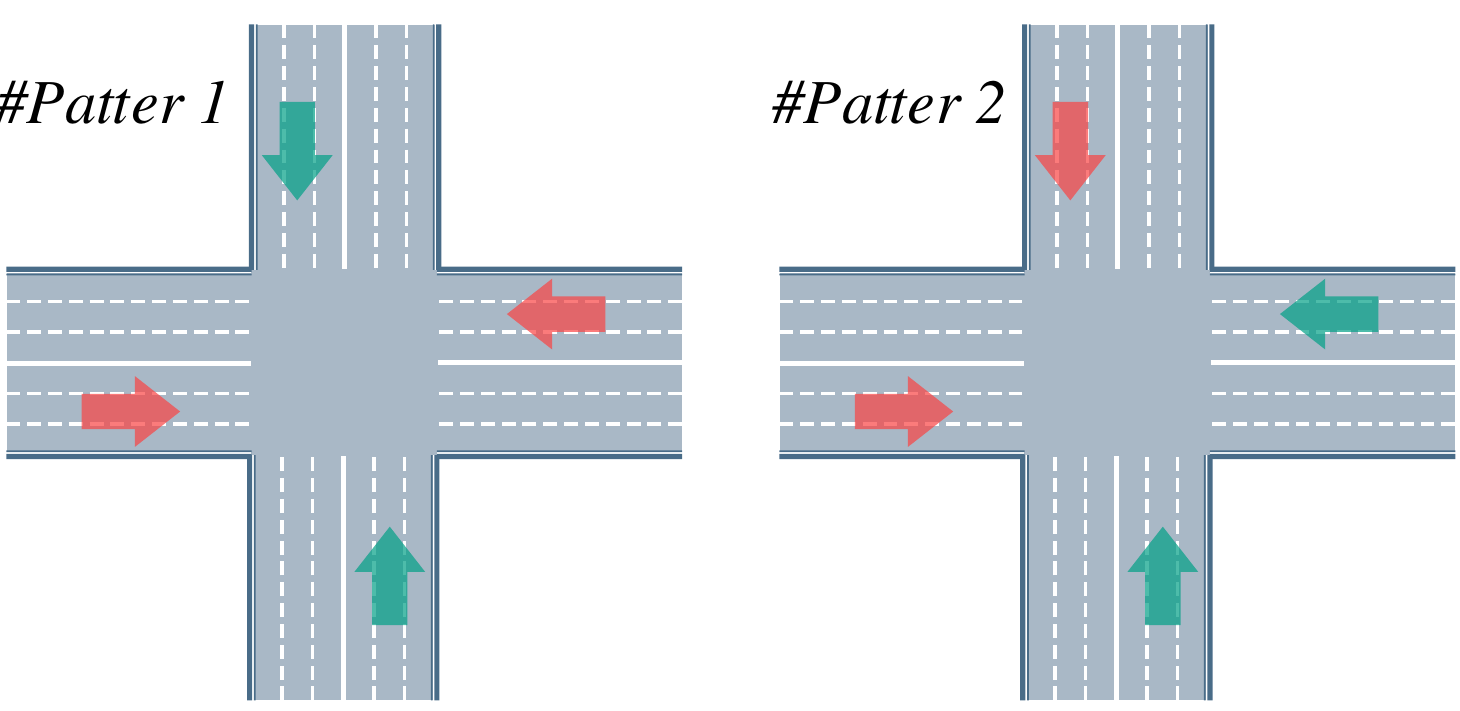}
\caption{Two typical heterogeneous traffic demand patterns. The traffic demand of the entrances indicated by the red arrows will be 1.5, 2.0, and 2.5 times that of the entrances indicated by green arrows. }
\label{fig12}
\end{figure}

\begin{table*}[!htb]
\footnotesize
\renewcommand{\arraystretch}{1.3}
\centering
\caption{Average delay under different heterogeneous traffic demands.}
\centering
\setlength{\tabcolsep}{4.5mm}{
\begin{tabular}{ccccc}
\hline\hline
\multicolumn{2}{c}{\multirow{2}{*}{Traffic demand}} & \multicolumn{3}{c}{Average delay (s)}                    \\ \cline{3-5}
\multicolumn{2}{c}{}                                & FIFO based algorithm & MCTS based algorithm & AlphaOrder \\ \hline
\multirow{3}{*}{\emph{Pattern 1}}         & 1.5        & 6.52                 & 4.71                 & 3.87       \\ \cline{2-5}
                                       & 2.0        & 5.87                 & 4.0                  & 3.60       \\ \cline{2-5}
                                       & 2.5        & 12.73                & 10.85                & 8.16       \\ \hline
\multirow{3}{*}{\emph{Pattern 2}}           & 1.5        & 20.62                & 17.88                & 12.81      \\ \cline{2-5}
                                       & 2.0        & 17.00                & 14.47                & 10.68      \\ \cline{2-5}
                                       & 2.5        & 22.27                & 19.84                & 14.23      \\
\hline\hline
\end{tabular}}
\label{table_3}
\end{table*}

\par Then, we explore two typical heterogeneous traffic demand patterns, as illustrated in Fig. \ref{fig12}. The heterogeneous traffic demands with 1.5, 2.0, and 2.5 times difference are simulated, respectively. We set the average arrival rate of all lanes as 300\emph{veh/(lane*h)} to keep the total arrival rate constant. Table \ref{table_3} lists the vehicles' average delay in different patterns. The results show that in heterogeneous traffic demand, AlphaOrder still outperforms FIFO and MCTS based algorithms and can obtain the passing order with a shorter delay-sum. Meanwhile, compared with \emph{\#Pattern 1}, the dominant traffic flows in \emph{\#Pattern 2} are conflictive, resulting in a greater average delay. However, the delay increment corresponding to AlphaOrder is smaller than that of FIFO and MCTS based algorithms. It suggests that AlphaOrder has excellent generalization for different traffic demands and can robustly deal with various traffic conditions.

\section{Conclusion}
\label{sec5}
\par In this work, we have proposed AlphaOrder, a novel cooperative driving algorithm based on deep learning and tree search. Unlike all the existing methods, we innovatively introduce the deep learning approach to solve the core problem of cooperative driving for CAV swarms, namely the passing order. AlphaOrder can learn the underlying experiences for producing promising passing orders and use them to address new scenarios. We show that AlphaOrder can find a near-optimal passing order for scenarios with an arbitrary number of CAVs, significantly superior to all exiting algorithms, thereby achieving the state-of-the-art performance. AlphaOrder also has excellent learning and transferring properties for efficient deployment to arbitrary new intersections. In addition, the implementation complexity of AlphaOrder is low enough to be applied on practical roadside unit, since the training/updating can be finished remotely on cloud.

\par AlphaOrder can address cooperative driving problems of unceasing CAV swarm flows in any, and especially larger, conflict areas. Moreover, we can also solve cooperative driving problems for CAV swarms in multi-conflict areas (e.g., traffic networks with various non-signalized intersections). Noticing that the key difficulty is still the right-of-way arrangements at conflicting areas, we can appropriately decompose the problem into small-scale sub-problems that address CAV cooperation within limited temporal-spatial areas and coordinate adjacent areas by specially designed information exchange. Thus, AlphaOrder can be conveniently used to address the cooperative driving problem in each small area \cite{zhang2022analysis, pei2021distributed}.

\par The generality of the core algorithm of AlphaOrder suggests that it can be applied to other problems concerning preemptive resource management. Such problems are widespread but still not effectively solved, e.g., resource management and scheduling for transportation systems \cite{Tian2016Dual, Khan2021Learning}, spatio-temporal resource management for swarm robotics systems \cite{yang2018grand}, and even more general resource management problems concerning natural resources, labor, and capital.

{\appendix[]
\begin{table}[!htb]
\footnotesize
\renewcommand{\arraystretch}{1.3}
\centering
\caption{The nomenclature list.}
\centering
\setlength{\tabcolsep}{4.5mm}{
\begin{tabular}{cp{5cm}}
\hline\hline
Symbol & Definition \\ \hline
 $p_\theta$ &  Pointer network, where $\theta$  denotes its weights.\\
 $b_\delta$ &  Critic network, where $\delta$   denotes its weights.\\
 $s$ &  A specific scenario, i.e., a problem instance.\\
 $N$ &  Number of vehicles in scenario $s$.\\
 $\{ V_1,\cdots, V_N \}$ &  Set of vehicle symbols in scenario $s$.\\
 $\bm{\pi}$ &  A passing order for scenario $s$.\\
 $\pi_k$&  The $k$-th element in $\bm{\pi}$.\\
 $Pi$ &  Set of all possible passing orders for scenario $s$.\\
 $J_{\text {Dealy }}\left(V_{i}, \bm{\pi} \mid s\right)$ &  Delay of \emph{Vehicle i} with the passing order $\bm{\pi}$ in scenario $s$.\\
 $f_{\text {Enforceable }}(\bm{\pi} \mid s)$ &  Judge the enforceability of $\bm{\pi}$ in scenario $s$.\\
 $J\left(\bm{\pi} \mid s\right)$&  Objective value of the formulated combinatorial optimization problem.\\
 $C$&  Penalty factor.\\
 $\bm{S}$ &  Input sequence of pointer network $p_\theta$.\\
 $\boldsymbol{S}^{\prime}$ &  Input sequence of critic network $b_\delta$.\\
 $ S_{V_i} $&  State representation of \emph{Vehicle i}.\\
 $D_{emb}$, $D$&  Vector dimensions.\\
 $Z^{time}_i$ &  Right-of-way state of conflict subzone \emph{i}.\\
 $M$ &  Number of groups.\\
 $B$ &  Batch size\\
 $\Psi$ &  Training dataset.\\
 $Q_i$ &  Value of node \emph{i} in a search tree.\\
 $T$,$T_i$&  Accumulated visit count of a node.\\
 $\lambda$, $\gamma$&  Hyper-parameters in MCTS algorithm.\\
\hline\hline
\end{tabular}}
\label{table_appendix_1}
\end{table}

}

\bibliographystyle{IEEEtran}
\bibliography{IEEEabrv,IEEEexample}
\begin{IEEEbiography}[{\includegraphics[width=1in,height=1.5in,clip,keepaspectratio]{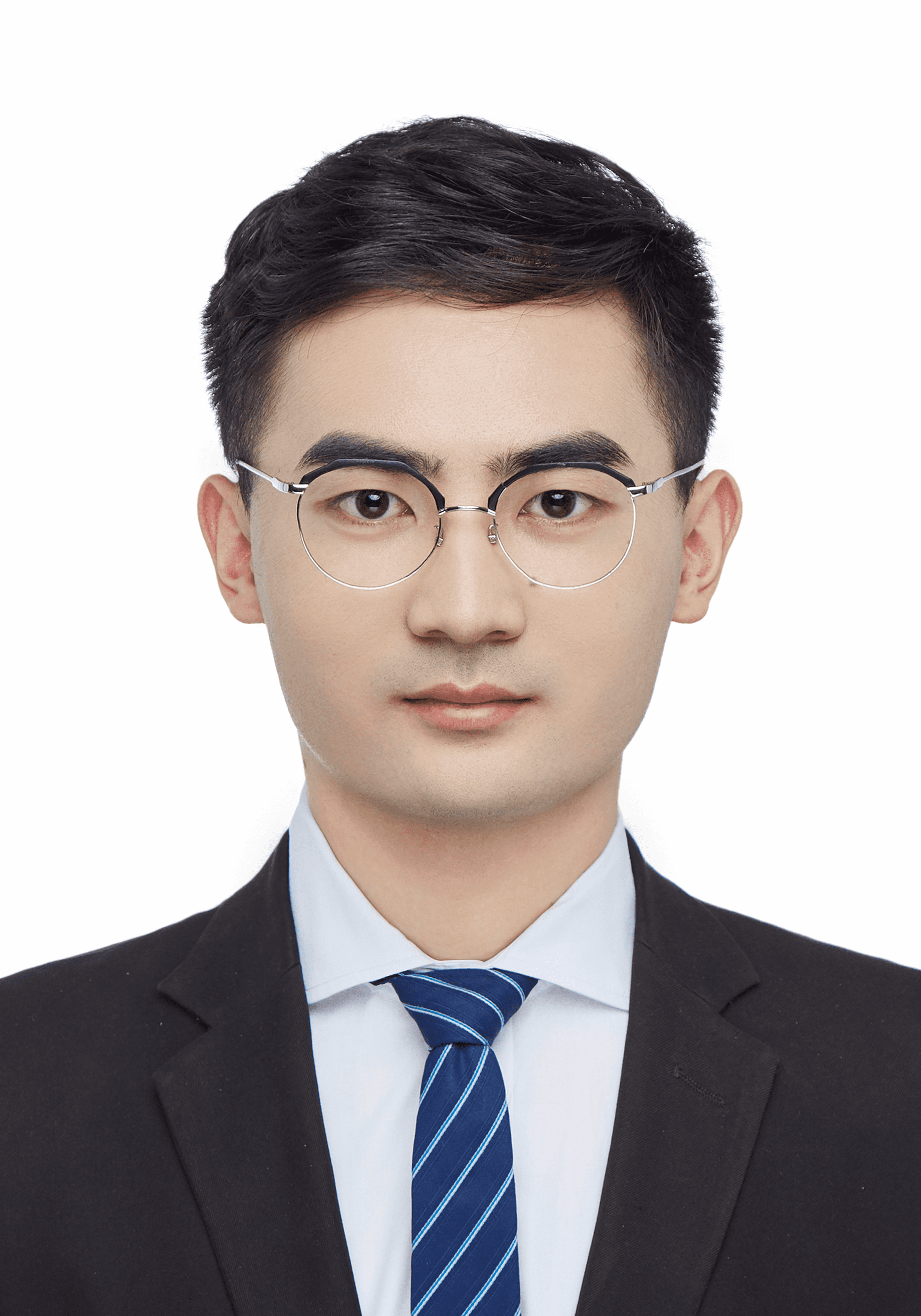}}]{Jiawei Zhang}
received the B.S. degree from Tsinghua University, Beijing, China, in 2020. He is currently pursuing the Ph.D. degree with the Department of Automation, Tsinghua University, Beijing, China. His research interests include autonomous driving, intelligent transportation systems, and deep reinforcement learning. He received the Best Student Paper Award at the 25th IEEE International Conference on Intelligent Transportation Systems.
\end{IEEEbiography}

\begin{IEEEbiography}[{\includegraphics[width=1in,height=1.5in,clip,keepaspectratio]{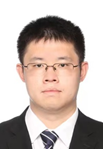}}]{Shen Li}
received the Ph.D. degree at the University of Wisconsin ¨C Madison in 2018. He is a research associate at Tsinghua University. His research is about Intelligent Transportation Systems (ITS), Architecture Design of CAVH System, Vehicle-infrastructure Cooperative Planning and Decision Method, Traffic Data Mining based on Cellular Data, and Traffic Operations and Management.
\end{IEEEbiography}

\begin{IEEEbiography}[{\includegraphics[width=1in,height=1.5in,clip,keepaspectratio]{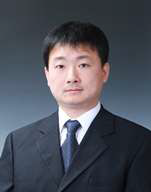}}]{Li Li}
(Fellow, IEEE) is currently a Professor with the Department of Automation, Tsinghua University, Beijing, China, working in the fields of artificial intelligence, intelligent control and sensing, intelligent transportation systems, and intelligent vehicles. He has published over 110 SCI-indexed international journal articles and over 70 international conference papers as a first/corresponding author. He is a member of the Editorial Advisory Board for the Transportation Research Part C: Emerging Technologies, and a member of the Editorial Board for the Transport Reviews and Acta Automatica Sinica. He also serves as an Associate Editor for IEEE Transactions on Intelligent Transportation Systems and IEEE Transactions on Intelligent Vehicles.
\end{IEEEbiography}

\end{document}